\documentclass[conference,compsoc]{ieeeconf}  % Comment this line out if you need a4paper

\IEEEoverridecommandlockouts                              % This command is only needed if 
\usepackage{cite}
\usepackage{xcolor,soul}
\usepackage{tcolorbox} 
\soulregister\cite7
\soulregister\citep7
\soulregister\citet7
\soulregister\ref7
\soulregister\pageref7

\usepackage{amssymb}
\usepackage{bm}
\usepackage{multirow}
\usepackage{soul}
\usepackage{svg}
\usepackage{float}

\usepackage{hyperref}

\hypersetup{
    colorlinks=true,
    linkcolor=black,
}

\usepackage{cuted}
\usepackage{capt-of}
\usepackage{caption}
\usepackage{amsmath}

\usepackage{blindtext}
\usepackage{tabularx}

\usepackage{amsmath}

\newtheorem{remark}{Remark}
\title{\LARGE \bf
UA-MPC: \underline{U}ncertainty-\underline{A}ware \underline{M}odel \underline{P}redictive \underline{C}ontrol for Motorized LiDAR Odometry}

\author{Jianping Li,~\IEEEmembership{Member,~IEEE}, Xinhang Xu, Jinxin Liu, Kun Cao, \\Shenghai Yuan,~\IEEEmembership{Member,~IEEE} and Lihua Xie,~\IEEEmembership{Fellow,~IEEE}}

\begin{document}
% \begin{linenumbers}
% \pagewiselinenumbers 
% \switchlinenumbers
\maketitle

\renewcommand{\thefootnote}{}
\footnotetext{J. Li, X. Xu, J. Liu, S. Yuan, and L. Xie are with School of Electrical and Electronic Engineering, Nanyang Technological University, 50 Nanyang Avenue, Singapore. K. Cao is with the Department of Control Science and Engineering, College of Electronics and Information Engineering, and Shanghai Research Institute for Intelligent Autonomous Systems, Tongji University, Shanghai, China. (E-mail: jianping.li@ntu.edu.sg, xinhang.xu@ntu.edu.sg, jinxin.liu@ntu.edu.sg, caokun@tongji.edu.cn, shyuan@ntu.edu.sg, elhxie@ntu.edu.sg)}

%%%%%%%%%%%%%%%%%%%%%%%%%%%%%%%%%%%%%%%%%%%%%%%%%%%%%%%%%%%%%%%%%%%%%%%%%%%%%%%%

\begin{abstract}

Accurate and comprehensive 3D sensing using LiDAR systems is crucial for various applications in photogrammetry and robotics, including facility inspection, Building Information Modeling (BIM), and robot navigation. Motorized LiDAR systems can expand the Field of View (FoV) without adding multiple scanners, but existing motorized LiDAR systems often rely on constant-speed motor control, leading to suboptimal performance in complex environments. To address this, we propose UA-MPC, an uncertainty-aware motor control strategy that balances scanning accuracy and efficiency. By predicting discrete observabilities of LiDAR Odometry (LO) through ray tracing and modeling their distribution with a surrogate function, UA-MPC efficiently optimizes motor speed control according to different scenes. Additionally, we develop a ROS-based realistic simulation environment for motorized LiDAR systems, enabling the evaluation of control strategies across diverse scenarios. Extensive experiments, conducted on both simulated and real-world scenarios, demonstrate that our method significantly improves odometry accuracy while preserving the scanning efficiency of motorized LiDAR systems. Specifically, it achieves over a 60\% reduction in positioning error with less than a 2\% decrease in efficiency compared to constant-speed control, offering a smarter and more effective solution for active 3D sensing tasks. The simulation environment for control motorized LiDAR is open-sourced at: \url{https://github.com/kafeiyin00/UA-MPC.git}.
    
\end{abstract}

%%%%%%%%%%%%%%%%%%%%%%%%%%%%%%%%%%%%%%%%%%%%%%%%%%%%%%%%%%%%%%%%%%%%%%%%%%%%%%%%
\section{Introduction }

Accurate and comprehensive 3D sensing of the environment using a LiDAR system is a cornerstone of advancements in photogrammetry \cite{li2023whu,li2024hcto,liao2023se} and robotics \cite{xu2022fast}. The high-resolution 3D point clouds generated by LiDAR systems enable a wide range of critical applications, including facility inspection \cite{cao2021distributed}, where detailed structural analysis is required; BIM \cite{kim2021development}, for precise 3D digitalization; and autonomous robot navigation \cite{zou2023patchaugnet, liu2015robotic}, where accurate environmental awareness is essential for path planning and obstacle avoidance \cite{cao2024learning}. However, a single LiDAR typically suffers from a limited Field of View (FoV), which can restrict its usability in tasks requiring full environmental coverage. To overcome this limitation, rotating the LiDAR with a motor to create a motorized LiDAR system significantly enhances the FoV without the need for additional LiDAR \cite{jiao2021robust}. This approach not only improves coverage but also offers advantages in terms of reduced weight, size, and system complexity, making it particularly well-suited for applications on lightweight and resource-constrained platforms, such as unmanned aerial vehicles (UAVs) \cite{hovermap2024,yang2022hierarchical} and mobile robots \cite{yuan2021low,zhen2017robust}. 

\begin{figure}
    \centering
    \includegraphics[width=0.5\textwidth]{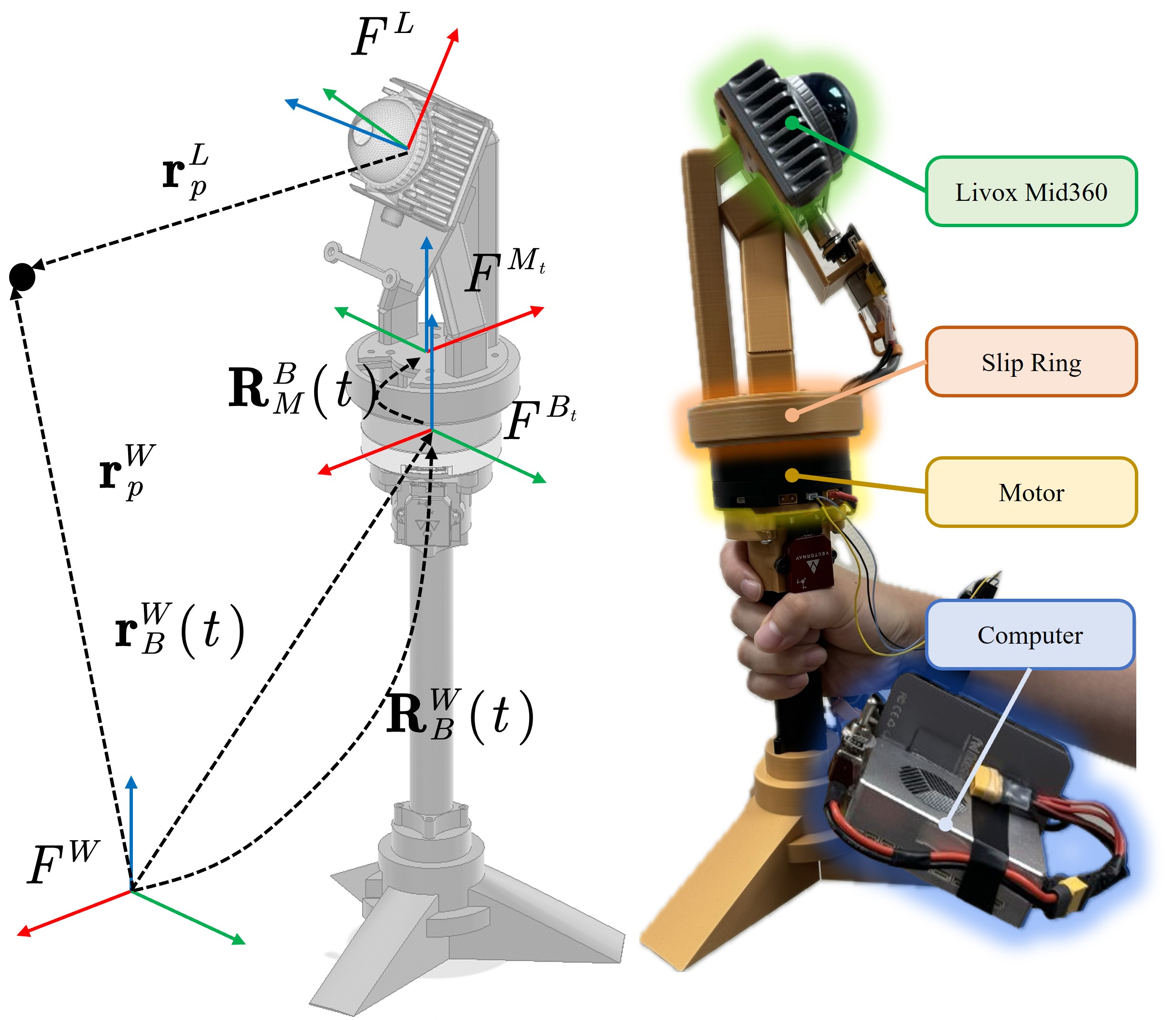}
    \caption{Coordinates and mechanical design of the proposed motorized LiDAR system. }
    \label{fig:coordinates}
    \vspace{-0.5cm}
\end{figure}

Thanks to the compact design and expansive FoV provided by rotating LiDAR systems equipped with motors, LO using motorized LiDAR systems has emerged as a key area of research over the past decade \cite{cong2024or,yuan2021low,zhen2017robust}. Early pioneering efforts \cite{zhang2014loam,kang2016full,alismail2015automatic} demonstrate the integration of 2D LiDAR with motors, enabling 3D sensing and the realization of 3D Simultaneous Localization and Mapping (SLAM) for robotic platforms. These foundational works highlight the potential of motorized LiDAR systems for dynamic and high-resolution environmental mapping. With advancements in LiDAR manufacturing technologies, the integration of 3D multi-beam LiDAR into motorized systems has further expanded their capabilities, as validated by recent studies \cite{ramezani2022wildcat}. To enhance SLAM accuracy on the motorized LiDAR system, several approaches have employed elastic mapping techniques \cite{park2021elasticity,park2018elastic}, which adaptively refine the map structure to achieve better accuracy. Additionally, innovative motor control strategies have been explored, such as the Lissajous rotating scheme \cite{karimi2021lola}, which utilizes two motors to generate optimized scanning patterns, significantly improving data quality for downstream SLAM processes. These existing studies collectively demonstrate the significant advancements made possible by motorized LiDAR systems. However, most existing studies on motorized LiDAR-based LO \cite{zhen2017robust,cong2024or,zhang2014loam} have employed constant-speed motor settings, determined largely by empirical experience rather than scene-specific optimization. 

Active LiDAR SLAM is gaining increasing attention from researchers due to its ability to dynamically adapt sensor behavior or vehicle paths to optimize performance in complex environments \cite{bai2024graph,chen2024lidar}. For our case, unlike traditional methods that rely on fixed scanning patterns, we want to incorporate adaptive control strategies, such as adjusting motor speeds, to achieve a balance between accuracy and efficiency. This adaptability allows the system to respond intelligently to varying environmental conditions and task-specific requirements, significantly enhancing its performance \cite{chen2020active}. 
This idea is similar to the work from \cite{cui2024alphalidar}, which proposes an adaptive control of a motorized LiDAR system considering the region of interest but without the accuracy of the LO. In challenging scenarios, such as feature-sparse, improper motor operation can lead to data degeneracy, adversely affecting both odometry accuracy and mapping reliability \cite{cong2024or}. 
To address these limitations, it is essential to explore methods for dynamically optimizing motor control to maximize system accuracy while maintaining efficiency. This necessitates a comprehensive understanding of the relationship between motor behavior, scene characteristics, and system outcomes. In light of these considerations, we identify two critical issues for motorized LiDAR systems that warrant further investigation:

\subsection{How to design a proper control strategy to balance the accuracy and efficiency?}

To the best of our knowledge, limited work has been done on developing effective motor control strategies for smarter motorized LiDAR systems. Intuitively, motorized LiDAR systems could improve LO accuracy by prioritizing areas with rich features, thereby enhancing feature association quality and overall localization precision. However, this targeted tracking strategy introduces potential trade-offs. Focusing on specific regions may lead to data holes in the point cloud, where parts of the environment are insufficiently scanned, negatively impacting global mapping consistency. Additionally, such strategies risk reducing scanning efficiency, as more time and resources are allocated to high-feature-density areas at the expense of overall coverage. Addressing these challenges requires a comprehensive approach that not only maximizes feature utilization but also ensures uniform rotating coverage and operational efficiency \cite{liu2023relative}. Developing smarter control strategies that dynamically balance these competing demands is crucial for advancing the capabilities of motorized LiDAR systems and unlocking their full potential in active SLAM applications.

\subsection{How to evaluate the different control strategies?}

Existing simulation environments (e.g., Gazebo \cite{koenig2004design}, Airsim \cite{shah2018airsim}, Isaac \cite{isaac_sim}) have played a crucial role in advancing research on LiDAR SLAM and motor control strategies. However, these platforms exhibit significant limitations, particularly in complex scenarios. A primary drawback is the considerable manual effort required to create detailed and realistic simulation environments. Designing complex scenes with diverse features, varying densities, and dynamic elements often demands substantial human labor, time, and expertise, limiting scalability and adaptability for broader applications.
MARSIM \cite{kong2023marsim} introduces a novel simulation approach by generating points directly from existing maps, serving as the foundation for our proposed simulation environment. Building upon this concept, we further develop tools to simulate motorized LiDAR scanning processes using data from existing SLAM datasets, streamlining the creation of diverse and realistic simulation scenarios.
Moreover, the existing simulation environments lack the fidelity required to effectively evaluate motor control strategies for motorized LiDAR systems. These platforms often fail to capture the nuanced characteristics of real-world LiDAR data. Consequently, they tend to produce overly optimistic results that do not reliably translate to real-world performance. This gap in realism poses challenges in accurately assessing the effectiveness of adaptive motor control strategies, particularly when balancing scanning efficiency and localization accuracy. Addressing these limitations is critical for developing smarter and more robust LiDAR-based systems.

Targeting the above two issues of the motorized LiDAR system, we propose UA-MPC for smarter motor control to balance the accuracy and efficiency of the motorized LiDAR system. The main contributions of the proposed method are as follows:

\begin{itemize}
\item To improve the scanning accuracies while maintaining the scanning efficiency, UA-MPC, an uncertainty-aware motor control strategy, considering the odometry accuracies and scanning efficiency is proposed for the motorized LiDAR system.
\item To solve the optimal motor control problem efficiently, discrete observabilities of the LiDAR odometry are first predicted using ray tracing, and then a surrogate function is used to model the distribution of the observabilities to speed up the optimization. 
\item We developed the first motorized LiDAR simulation environment under ROS with the adaptation to the existing SLAM dataset. The simulation environment can be easily used to evaluate different motor control strategies for motorized LiDAR in various scenes. We also demonstrate the effectiveness of the proposed system using an in-house hardware system as shown in Fig. \ref{fig:coordinates}.
\end{itemize}

\section{Uncertainty-Aware Model Predictive Control (UA-MPC)}

\subsection{Notations and Hardware System}

\subsubsection{Coordinates and poses} 

\begin{figure}
    \centering
    \includegraphics[width=0.48\textwidth]{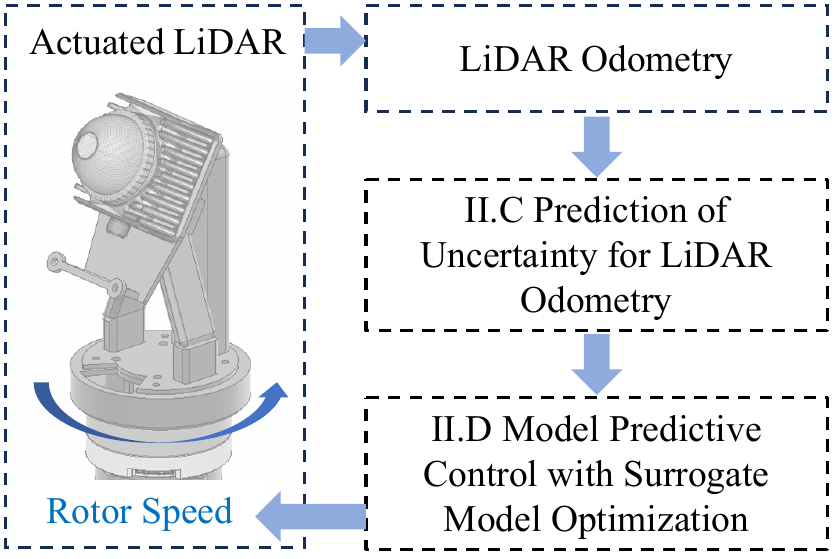}
    \caption{Workflow of \underline{U}ncertainty-\underline{A}ware \underline{M}odel \underline{P}redictive \underline{C}ontrol (UA-MPC). }
    \label{fig:workflow}
    \vspace{-0.5cm}
\end{figure}

In this paper, we use italic, bold lowercase, and bold uppercase letters to represent scalars, vectors, and matrices, respectively. As shown in Fig.\ref{fig:coordinates}, four coordinate frames are used in our proposed method, namely, the LiDAR frame $F^{L}$, the motor frame $F^{M_t}$  at time $t$, the base frame $F^{B_t}$ at time $t$, and the world frame $F^{W}$. We denote a point observed by the LiDAR in the sensor frame as $\mathbf{r}^{L}_p \in \mathbb{R}^3$, which is represented as $\mathbf{r}^{W}_p \in \mathbb{R}^3$ in the world frame. The transformation between the $\mathbf{r}^{L}_p$ and $\mathbf{r}^{W}_p$ is written as follow:

\begin{equation}
    \mathbf{r}^{W}_p = \mathbf{R}^W_B(t)\left(\mathbf{R}^B_M(t)\left(\mathbf{R}^M_L \mathbf{r}^{L}_p + \mathbf{r}^M_L \right) \right)+\mathbf{r}^W_B(t), \label{eq:coordinate_projection}
\end{equation}
where $\mathbf{R}^M_L \in \mathbb{SO}^3$ and $\mathbf{r}^M_L \in \mathbb{R}^3$ are the rotation and translation calibration parameters between the LiDAR and motor. $\mathbf{R}^B_M(t)\in \mathbb{SO}^3$ is the time-varying rotation matrix obtained by the angular encoder of the motor. $\mathbf{R}^W_B(t) \in \mathbb{SO}^3$ and $\mathbf{r}^W_B(t)\in \mathbb{R}^3$ are the rotation and translation to be estimated by the LiDAR odometry system. In this work, we utilize our previous work, I2EKF-LO \cite{yu2024i2ekf} as the LiDAR odometry algorithm.

\remark {According to the geometry projection in Eq.~\eqref{eq:coordinate_projection}, we only consider the odometry of the base frame ( $\mathbf{R}^W_B(t)$ and $\mathbf{r}^W_B(t)$) to avoid estimating the high dynamic rotation motion of the sensor. Thus, we can still easily use the system dynamic of the vehicle to predict the initial guess of the system pose, which is essential for the convergence of the Extended Kalman Filter (EKF) \cite{yu2024i2ekf}.}

In this work, we focus on the control of motor state $\mathbf{R}^B_M(t)$ and simplify it as $\mathbf{R}(t)$. As the motor in the motorized system only has one degree of freedom along the z-axis with the angle $\theta(t)$, the transformation between the rotation matrix $\mathbf{R}(t)$ and the angle $\theta(t)$ is written as:
\begin{align}
    \mathbf{R}(t) =
    \begin{bmatrix}
    \cos\theta(t) & -\sin\theta(t) & 0 \\
    \sin\theta(t) & \cos\theta(t) & 0 \\
    0 & 0 & 1
    \end{bmatrix} \in \mathbb{R}^{3 \times 3}.
\end{align}

In practice, discrete times $\{t_i,i=1,...,N\}$ with identical time interval $\Delta t$ are considered instead of the continuous time. The state transformation of discrete-time states $\mathbf{\Theta} = \{\theta_i,i=1,...,N\}$ for motor angle are written as follow:
\begin{align}
    \theta_{i+1} = \theta_i + \omega_i \Delta t,
\end{align}
where $\theta_i$ is the simplification of $\theta (t_i)$. $\omega_i$ is the motor speed at time instance $t_i$.

\subsection{Problem and System Overview}

\begin{figure}
    \centering
    \includegraphics[width=0.48\textwidth]{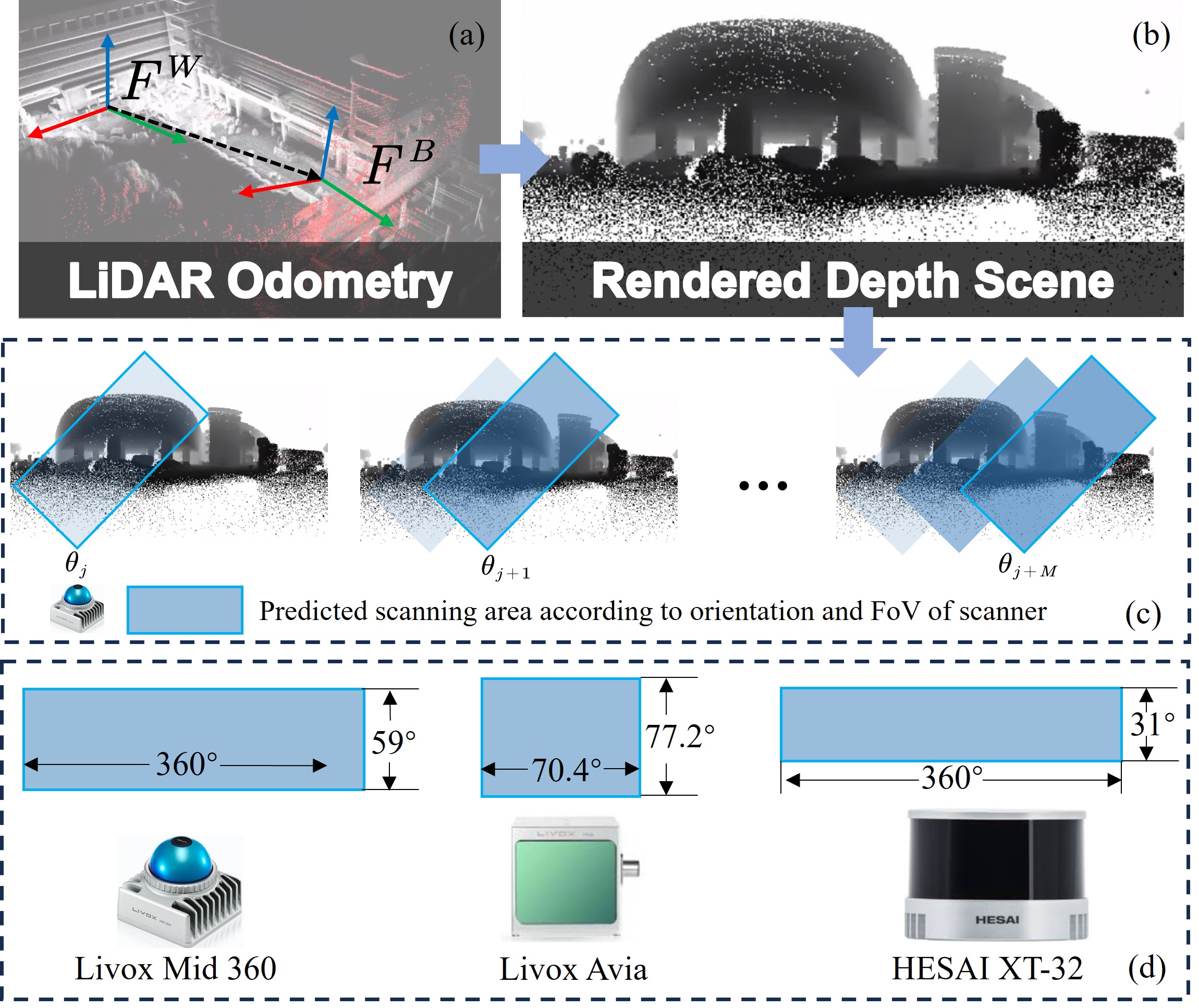}
    \caption{Prediction of the uncertainty of the LiDAR Odometry (LO) at different observing orientations controlled by the motor. (a) Taking the local map from LO as the input. (b) Rendering the panoramic depth map using the local map. (c) Sampling of the LiDAR measurements at specific orientation to calculate $U(\theta_j)$. (d) The typical laser scanner with different vertical and horizontal Field of View (FoV).}
    \label{fig:prediction}
    \vspace{-0.5cm}
\end{figure}

To the best of our knowledge, most of the existing motorized LiDAR systems \cite{zhen2017robust,cong2024or,zhang2014loam} directly set the motor at a constant speed according to their experience without considering the optimal control of the motor according to different scenes. Hence, the existing motorized LiDAR systems still face degeneracy problem in challenging environments \cite{cong2024or}. Targeting a smarter control of motor states $\mathbf{\Theta}$ to achieve better odometry accuracies while maintaining scanning efficiency, we propose the UA-MPC illustrated in Fig.~\ref{fig:workflow}. The LiDAR data stream from the motorized LiDAR system is fed to a LiDAR Odometry (I2EKF \cite{yu2024i2ekf} is used in our system). 

Then for the current time $t_j$, and the corresponding control horizon $\mathbf{\Omega}_j = \{\omega_i|i=j-M,j-M+1,...,j+M\}$, we want to achieve the balance between odometry accuracies and scanning efficiency by minimizing the function $F(\mathbf{\Omega}_j)$:

\begin{subequations}
\begin{align}
    \min_{\mathbf{\Omega}_j} F(\mathbf{\Omega}_j) = & \alpha \sum^{j+M}_{i=j-M} \Vert U(\theta_i)\Vert ^2 + \beta \sum^{j+M-1}_{i=j-M} \Vert E(\omega_i) \Vert ^2,  \label{eq:problem_main}\\
    E(\omega_i) = & \omega_i - \omega_{pre}, \label{eq:problem_energy}\\
    \theta_{i+1} = & \theta_i + \omega_i \Delta t. \label{eq:problem_theta}
\end{align}
\end{subequations}

The function $U(\theta_i)$ evaluates the uncertainty of the LiDAR odometry at the specific orientation $\theta_i$, which is a black box function related to the nearby scene and will be explained in section \ref{sec:prediction}. Then function $E(\omega_i)$ evaluates the efficiencies of the control horizon by comparing the motor speed with the preset speed $\omega_{pre}$. Intuitively, the function $E(\omega_i)$ guarantees the efficiency because the function $E(\omega_i)$ tends to focus on a fixed direction to increase the feature association and lead to zero speed. To solve the optimization problem with the black-blox function $U$, a surrogate model optimization is used and detailed in section \ref{sec:optimization}. We balance the two factors by properly setting the weights $\alpha$ and $\beta$. After getting the motor speeds in the control horizon $\mathbf{\Omega}_j$, we only execute one step $\mathbf{\omega}_j$. The details are described in the following sections. 

\subsection{Prediction of Uncertainty for LiDAR Odometry }\label{sec:prediction}

As we consider the accuracies of the LiDAR Odometry at different observing orientations of the LiDAR, we predict the uncertainty of LiDAR Odometry using the function $U$. The core idea of the uncertainty prediction function $U$ is taking the local point cloud map from the LO as input, and then simulating the sampled LiDAR measurements using ray-tracing in the future observing orientations $\theta_j$ as shown in Fig.~\ref{fig:prediction}. 

More specifically, setting the origin of base frame $F^{B_{ti}}$ as the projection center, the local points from the LO (Fig. \ref{fig:prediction} (a)) are projected onto the panoramic coordinates to obtain the panoramic depth map (Fig. \ref{fig:prediction} (b)) using Eq.~\eqref{eq:pano_projection}:
\begin{align}
    \left[ \begin{array}{c}
	u\\
	v\\
\end{array} \right] =\left[ \begin{array}{c}
	\frac{\mathrm{arc}\tan\left( y,x \right)}{2\pi} W\\
	\frac{\mathrm{arc}\sin \left( z/\sqrt{x^2+y^2+z^2} \right)}{\pi} H\\
\end{array} \right],\label{eq:pano_projection}
\end{align}
where $[x,y,z]$ denotes a point in current base frame $F^{B_{ti}}$. $[u,v]$ denotes the panoramic coordinates. $[W, H]$ denotes the width and height of the image. Then the future observation of the LiDAR at different motor angles can be sampled according to the Field of View (FoV) of LiDAR as illustrated in Fig. \ref{fig:prediction} (c). Three typical scanners with different FoV are illustrated in Fig. \ref{fig:prediction} (d). For simplification and efficiency, we only sampled the LiDAR measurements with uniform patterns (5 degrees for horizontal and vertical directions) without considering the non-repetitive scanning mode of the solid-stare scanners.

\remark {In the rendering steps described above, the panoramic depth image is rendered only once for the following uncertainty prediction assuming the body frame is static in a short time duration to save computational resources. An alternative way is using the constant velocity model for the base frame, and rendering the depth image several times. However, the proposed method is mainly used for edge computing units on low-speed handheld devices without GPU, we only render the panoramic depth image once.}

The rendered LiDAR measurements at the specific motor angle $\theta_i$ are then used to analyze the uncertainty of the state estimation in the LO at time $t_i$. The rendered $K$ LiDAR measurements in the base frame are denoted as $\{\mathbf{p}_k,k=0,...,K\}$. For the $k$-th point $\mathbf{p}_k$, its corresponding point in the world frame and normal vector are denoted as $\mathbf{P}_k$ and $\mathbf{n}_k$. For the point-to-plane measurements used for LO, the residual $\epsilon_k$ for the point $\mathbf{p}_k$ is calculated using Eq. \eqref{eq:residual}:

\begin{align}
\epsilon_k=\mathbf{n}_k(\hat{\mathbf{R}}^W_{B_i} \mathbf{p}_k + \hat{\mathbf{r}}^W_{B_i} - \mathbf{P}_k), \label{eq:residual}
\end{align}
where $\hat{\mathbf{R}}^W_{B_i}$ and $\hat{\mathbf{r}}^W_{B_i}$ are the initial guess for the pose of the base frame, which needs to be optimized in the least square estimation in LO. The Jacobian of the residual with respect to pose can be calculated using Eq. \eqref{eq:jacobian}:

\begin{equation}
\begin{aligned}
\mathbf{J}_k &= [(\partial\epsilon_k /\partial \hat{\mathbf{R}}^W_{B_i})^\top,(\partial\epsilon_k /\partial \hat{\mathbf{r}}^W_{B_i})^\top]^\top \\
&= [[(\hat{\mathbf{R}}^W_{B_i}\mathbf{p}_k]_{\times}\mathbf{n}_k)^\top,\mathbf{n}^\top_k]^\top. \label{eq:jacobian}
\end{aligned}
\end{equation}

According to the optimal design theory \cite{pukelsheim2006optimal}, A-optimal, the trace of the inverse of the information matrix, is used to evaluate the uncertainty of the accuracies of the LO at the motor angle $\theta_i$: 

\begin{subequations}
\begin{align}
U(\theta_i) &= \mathrm{Trace}(\mathbf{\varLambda}^{-1}_i),\label{eq:uncertainty}\\ 
\mathbf{\varLambda}_i &= \Sigma_{k=0}^{K} \mathbf{J}_k \mathbf{J}^{\top}_k, 
\end{align}
\end{subequations}
where $\mathbf{\varLambda}_i$ is the information matrix for the rendered points at rotation angle $\theta_i$.

\subsection{Model Predictive Control with Surrogate Model Optimization}\label{sec:optimization}

\begin{figure}
    \centering
    \includegraphics[width=0.48\textwidth]{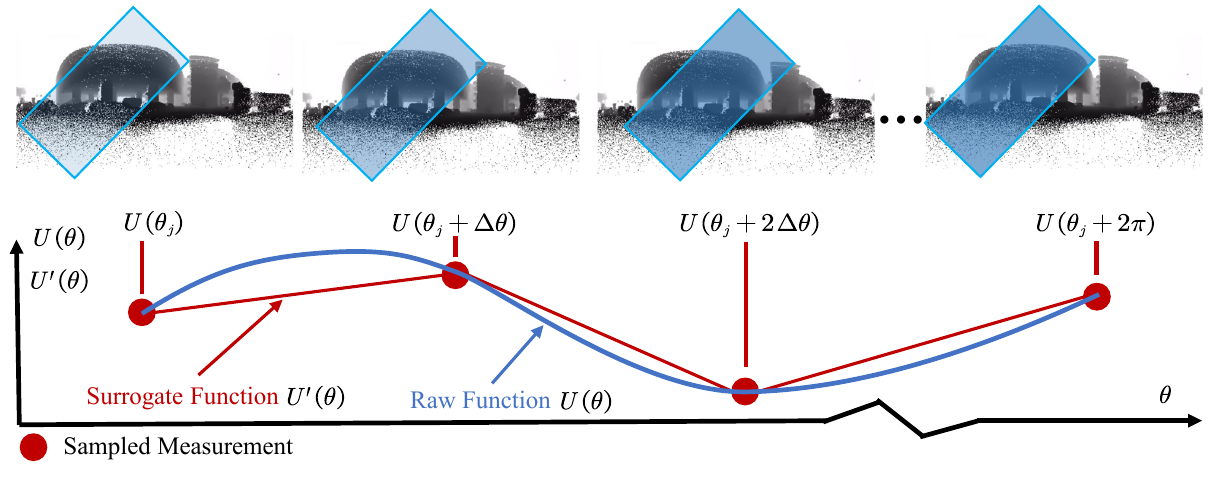}
    \caption{Surrogate function $U^\prime$, a piecewise linear function, is used for approximating the raw uncertainty function $U$. }
    \label{fig:surrogate}
    \vspace{-0.5cm}
\end{figure}

The uncertainty function $U$ is a black box function, as it is related to the rendering of the environment using the local map from LO. Moreover, the calculation of the uncertainty function $U$ involves rendering and point cloud sampling, which makes the numerical differential-based optimization method for the black-box function hard to implement in the edge computing unit. Fortunately, the points from successive motor steps share a lot of overlaps, which makes the uncertainty function $U$ changes slowly along the motor angle $\theta_j$ in the control horizon. Thus we use a piecewise linear surrogate function $U^\prime$ to approximate $U$ in the model predictive control to achieve a fast and effective control as shown in Fig. \ref{fig:surrogate}. 

More specifically, with the interval of $\Delta \theta$, $2 \pi/\Delta \theta$ samples are collected for $U$ in the current control horizon $\mathbf{\Omega}_j$. The samples are denoted as Eq. \eqref{eq:samples}:

\begin{align}
\{U_s=U(\theta_j+s\Delta \theta),s=0,1,...,2 \pi/\Delta \theta -1 \}. \label{eq:samples}
\end{align}
Then the surrogate function $U^\prime$ is constructed using the sample $\{U_s\}$. The $U(\theta_i)$ is approximated as follows:

\begin{equation}
\begin{aligned}
U(\theta_i) \approx U^\prime(\theta_i)  =  &(1 - \frac{\theta_i}{\Delta \theta} + 
\lfloor\frac{\theta_i}{\Delta \theta}\rfloor) U_{\lfloor\frac{\theta_i}{\Delta \theta}\rfloor} + \\
&(\frac{\theta_i}{\Delta \theta} - \lfloor\frac{\theta_i}{\Delta \theta}\rfloor) U_{\lfloor\frac{\theta_i}{\Delta \theta}\rfloor + 1} , \label{eq:approx_u}
\end{aligned}
\end{equation}
The function $U(\theta_i)$'s Jacobian with respect to $\theta_i$ is written as: 
\begin{equation}
\begin{aligned}
\partial  U(\theta_i)/ \partial \theta_i &\approx \partial  U^\prime(\theta_i)/ \partial \theta_i\\
&=(U_{\lfloor\frac{\theta_i}{\Delta \theta}\rfloor + 1} - U_{\lfloor\frac{\theta_i}{\Delta \theta}\rfloor} )/ \Delta \theta. \label{eq:approx_partial_u}
\end{aligned}
\end{equation}
With the approximation of $U(\theta_i)$ using Eq. \eqref{eq:approx_u} and \eqref{eq:approx_partial_u}, the optimization function $F(\mathbf{\Omega}_j)$ in Eq. (\ref{eq:problem_main}) can be solved using MPC solver. In practice, we first initialize all the control steps in $\Omega_j$ as the preset speed $\omega_{pre}$, then iteratively optimize all control steps.

\section{Realistic Simulation Environment for Motorized LiDAR System}

\begin{figure}
    \centering
    \includegraphics[width=0.48\textwidth]{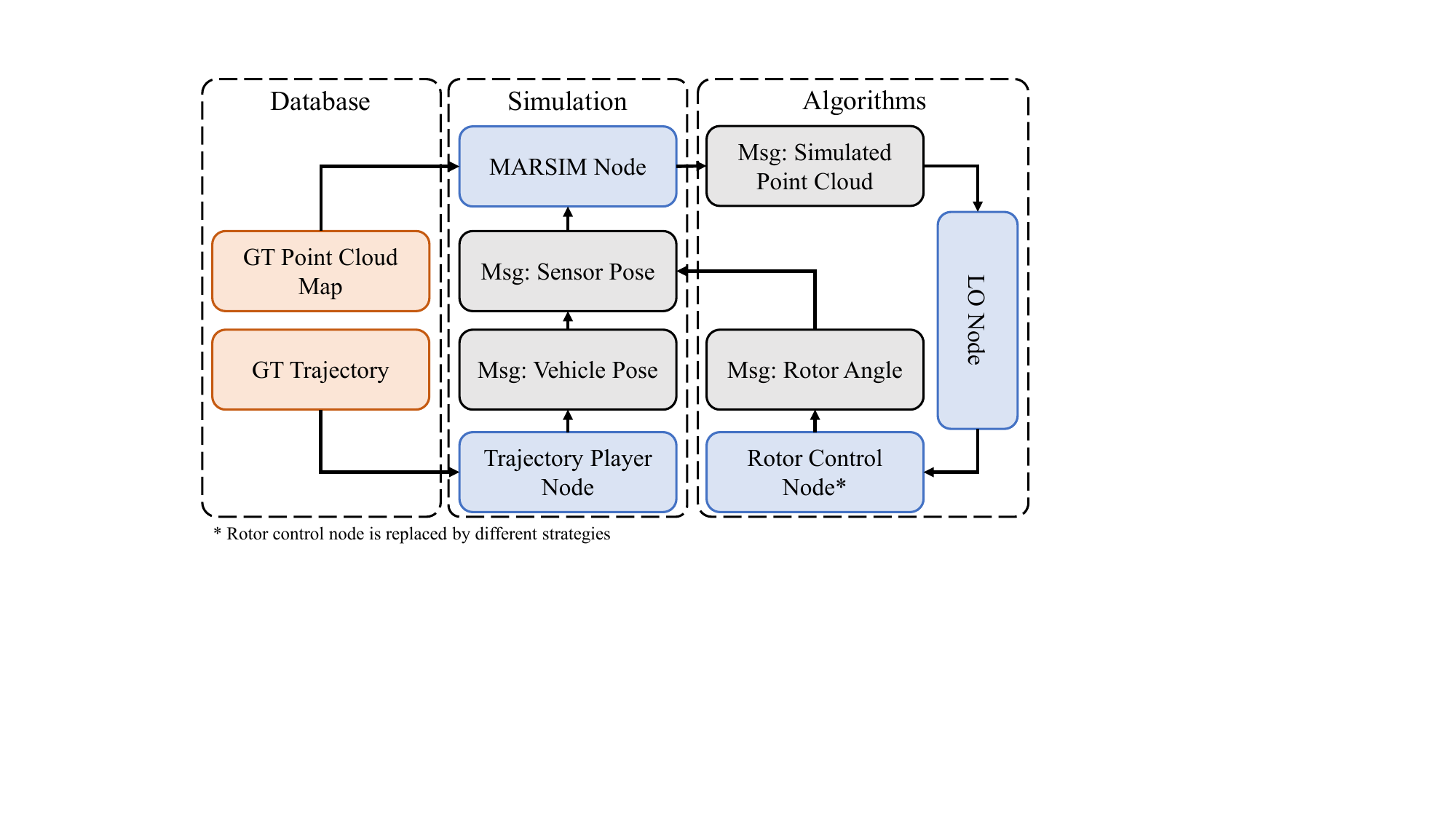}
    \caption{Simulation environment for motorized LiDAR system based on MARSIM and ROS.}
    \label{fig:simu_env}
    \vspace{-0.5cm}
\end{figure}

\begin{figure*}
    \centering
    \includegraphics[width=0.8\textwidth]{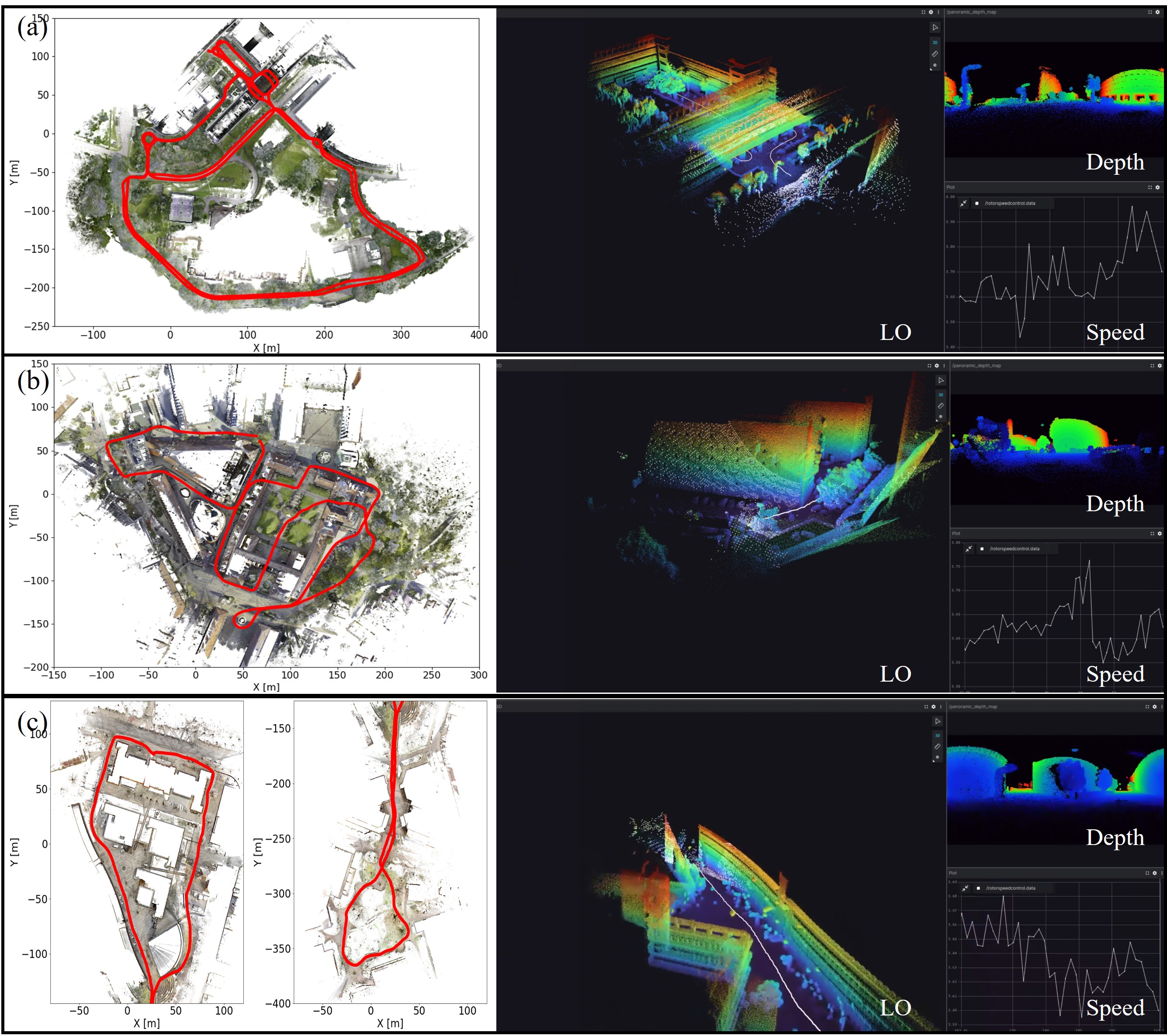}
    \caption{Simulation experiments using the proposed UA-MPC on different scenes in the MCD dataset. (a) NTU Campus; (b) KTH Campus; (c) TUHH Campus.}
    \label{fig:simulation_experiments}
    \vspace{-0.5cm}
\end{figure*}

To facilitate the evaluation of various motor control strategies within a realistic simulation environment, we develop a dedicated motorized LiDAR simulation tool built on MARSIM \cite{kong2023marsim} and Robot Operating System (ROS), as depicted in Fig. \ref{fig:simu_env}. This simulation framework takes a Ground Truth (GT) trajectory and a corresponding GT point cloud map as inputs, both of which can be sourced from most existing SLAM datasets. In this study, we utilize the Multi-Campus Dataset (MCD) \cite{nguyen2024mcd} to construct the simulation database, leveraging its diverse and complex environmental representations. The simulation process begins with a trajectory player node that replays the GT trajectory. The published GT odometry data is then combined with the rotor angle from the designed rotor control node for the sensor pose. The sensor pose is fed to MARSIM and laser ray tracing to generate the simulated LiDAR measurements. This setup not only ensures high fidelity in simulating real-world sensor behaviors but also enables researchers to replace both the LO node and the motor control node, allowing for seamless validation and comparison of different algorithms. The flexibility and extensibility of this simulation environment make it a powerful tool for testing and optimizing motor control strategies under varied conditions.

\section{Experiments}

\subsection{Implementation Details}

We implemented UA-MPC using C++ and ROS. Based on our empirical observations, the reference rotating speed, $\omega_{\text{pre}}$, is set to $3.6~\text{rad/s}$. The magnitude of the uncertainty factor $U(\theta_i)$ differs from the angular velocity factor $E(\omega_i)$ by approximately three orders of magnitude. To balance their contributions to the system, we assign weights of $\alpha = 1000$ and $\beta = 1$, ensuring that both factors influence the control strategy equally. For the simulation experiments, evaluations were performed on a computer equipped with an Intel i9-10940 CPU and an NVIDIA GeForce RTX 4090 GPU. For real-world testing, UA-MPC was deployed and tested on an NVIDIA Orin-NX, demonstrating its applicability and efficiency across different hardware platforms.

\subsection{Evaluation Metrics}

We evaluate the different control strategies for the motorized LiDAR system from two perspectives: (1) the LO accuracy and (2) the scanning efficiency. The Absolute Translation Error (ATE) is used to measure the LO accuracy:

\begin{equation}
\text{ATE} = \sqrt{\frac{1}{N} \sum_{i=1}^{N} \|\mathbf{r}_i^{\text{est}} - \mathbf{r}_i^{\text{gt}}\|^2},
\end{equation}
where $\mathbf{r}_i^{\text{est}}$ and $\mathbf{r}_i^{\text{gt}}$ denote the estimated and ground truth positions, respectively, and $N$ represents the total number of pose estimates.

For scanning efficiency, we calculate the number of voxels (each with a size of $0.5~\text{m}$) scanned by the system every 5 seconds and compute the average value to evaluate the completeness (CMPLT):

\begin{equation}
\text{CMPLT} = \frac{1}{Q} \sum_{q=1}^{Q} v_q,
\end{equation}
where $v_q$ is the number of scanned voxels during the $q$-th scan period, and $Q$ is the total number of scan periods. This dual evaluation framework enables a comprehensive assessment of both the accuracy and the efficiency of different control strategies.

\subsection{Evaluation on Simulation Environment}

To evaluate the performance of the proposed UA-MPC, we compare it with two alternative motor control strategies commonly used in existing systems: constant-speed control (fixed at $3.6~\text{rad/s}$) and zero-speed control. The simulation experiments are conducted on the MCD dataset, as illustrated in Fig. \ref{fig:simulation_experiments}. During the experiments, we simultaneously visualize the LO, the panoramic depth map used for motor control, and the motor speed command. A video recording of the simulation experiments is available at \url{https://www.youtube.com/watch?v=zkbm0Tkp-PM}. From visual inspection, the proposed UA-MPC dynamically adjusts the rotation speed in response to unbalanced features in the panoramic depth map. Specifically, UA-MPC focuses on areas with rich features while maintaining a balance between scanning efficiency and LO accuracy, demonstrating its adaptability and effectiveness in diverse scenarios.

\begin{figure}
    \centering
    \includegraphics[width=0.48\textwidth]{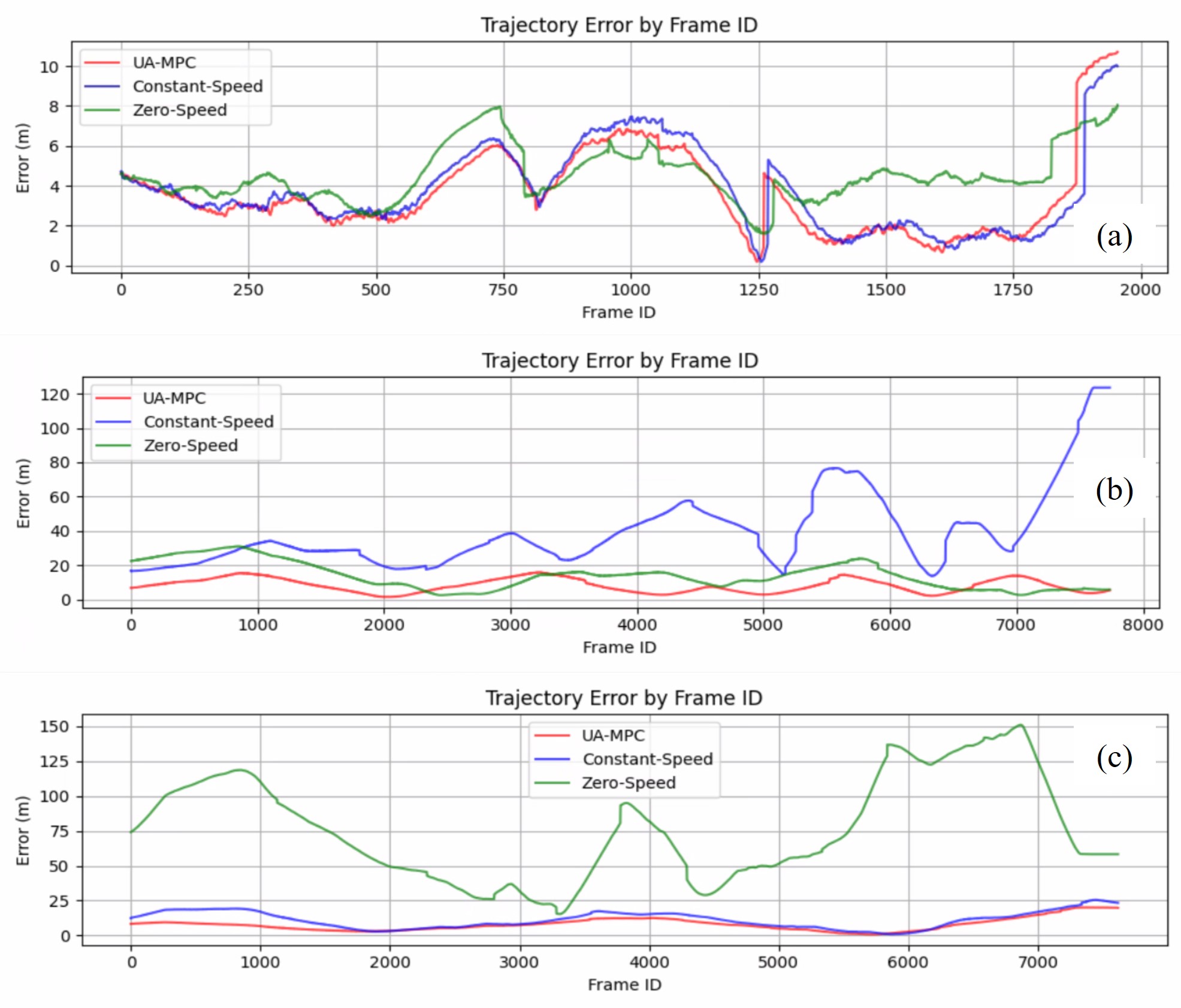}
    \caption{Absolute Translation Error (ATE) of different methods on the simulation dataset. (a) NTU Campus; (b) KTH Campus; (c) TUHH Campus.}
    \label{fig:simu_ate}
\end{figure}

The comparisons of UA-MPC, constant-speed control, and zero-speed control are presented in Table \ref{tab:eva_simu}. The results demonstrate that the proposed UA-MPC achieves the best LO accuracy across all three sites of the MCD dataset. This improvement can be attributed to UA-MPC's ability to dynamically adjust the motor speed, focusing on areas with rich features, which enhances feature tracking and reduces the likelihood of tracking failures. In contrast, the constant-speed and zero-speed control strategies are not adaptive, leading to occasional feature tracking failures. The ATE values for each control strategy are visualized in Fig. \ref{fig:simu_ate}, highlighting the differences in localization performance. Notably, the constant-speed control exhibits significant drift in the MCD-KTH site, while the zero-speed control suffers from drift in the MCD-TUHH site. These results emphasize the limitations of fixed or static motor speed strategies in complex environments. Furthermore, Table \ref{tab:eva_simu} indicates that UA-MPC achieves scanning completeness (CMPLT) comparable to constant-speed control across all MCD sites. This demonstrates that UA-MPC effectively balances localization accuracy and scanning efficiency, providing a robust and adaptable solution for motorized LiDAR systems in diverse scenarios. Overall, compared to the constant-speed control, the ATE is improved by 60\% using UA-PMC, while the CMPLT is reduced by less than 2\%.

\begin{table}[]
\caption{Evaluation on Simulation Dataset \label{tab:eva_simu}}
\centering
\scalebox{0.8}{
\begin{tabular}{c|cc|cc|cc}
\hline
               & \multicolumn{2}{c|}{NTU} & \multicolumn{2}{c|}{KTH} & \multicolumn{2}{c}{TUHH} \\
               & ATE (m)    & CMPLT      & ATE (m)    & CMPLT       & ATE (m)    & CMPLT      \\ \hline
UA-MPC         & 4.55       & 62068       &  9.01      &  32916      & 9.39       & 37686       \\
Constant-Speed & 4.96       & 69237       &  45.03     &  33329      & 13.21      & 37779       \\
Zero-Speed     & 4.71       & 51977       &  15.3      &  22529      & 84.64      & 24737       \\ \hline
\end{tabular}
}
\end{table}

\subsection{Evaluation on In-house Handheld motorized LiDAR System}

\begin{figure}
    \centering
    \includegraphics[width=0.48\textwidth]{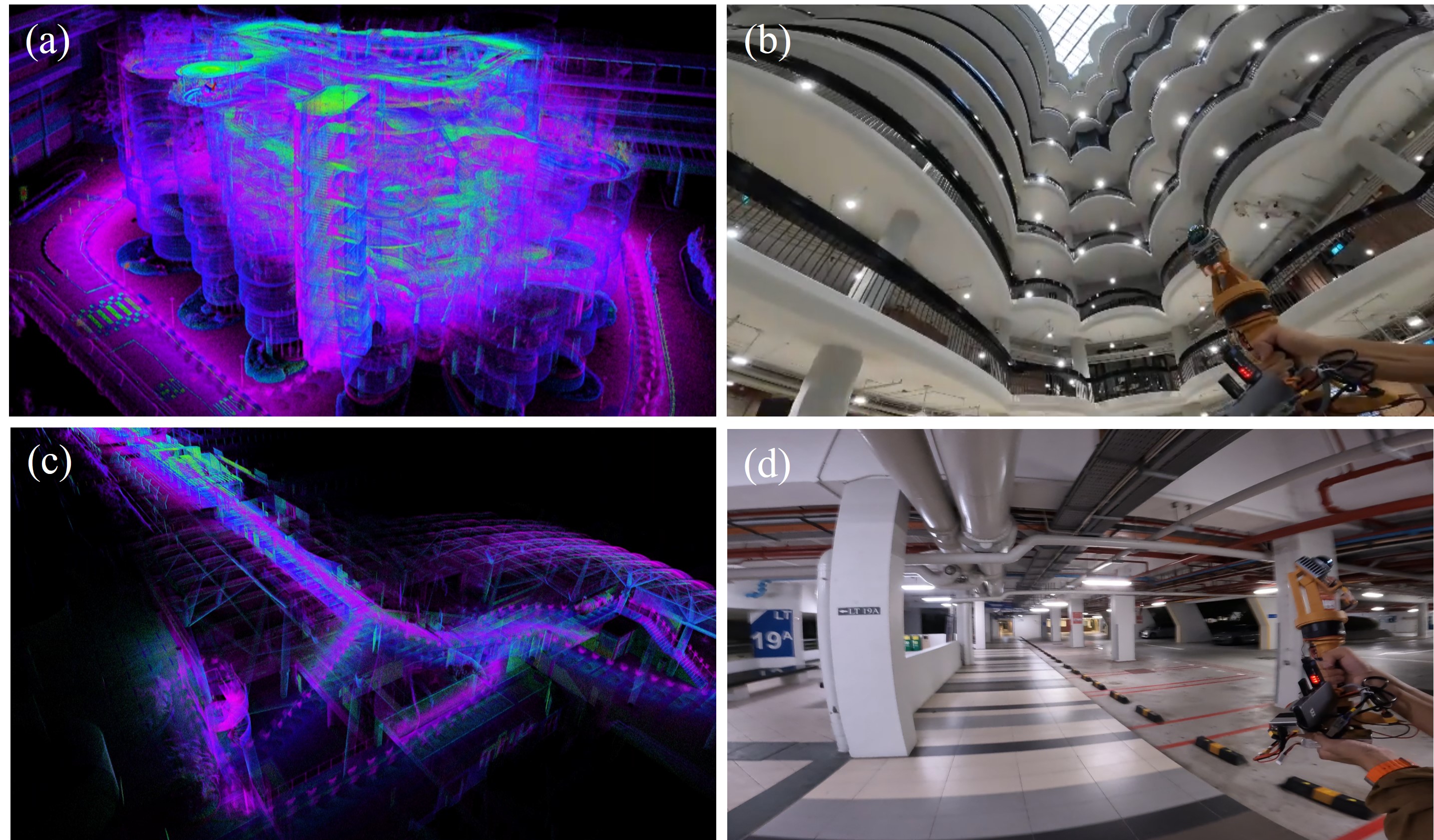}
    \caption{Experiments on in-house handheld motorized LiDAR system. (a) The final point clouds for the NTU Hive. (b) Scanning using the handheld system in NTU Hive. (d) The final point clouds for the NTU Spine. (e) Scanning using the handheld system in NTU Spine.}
    \label{fig:real_experiments}
    \vspace{-0.5cm}
\end{figure}

To further validate the effectiveness of the proposed UA-MPC, we integrated it into an in-house handheld motorized LiDAR system, with the principal mechanical drawing illustrated in Fig. \ref{fig:coordinates}. The system was evaluated in two challenging scenes at NTU, as depicted in Fig. \ref{fig:real_experiments}. The first test site, NTU Hive, is a multi-level complex building with intricate architectural features, as shown in Fig. \ref{fig:real_experiments}(b). The scanning trajectory covered both outdoor and indoor environments, including narrow staircases, as we navigated from the first floor to the top. This path introduced numerous degenerate scenarios, such as feature-sparse staircases, which are particularly challenging for existing mapping systems. Despite these complexities, our system successfully captured an accurate 3D digitalization of the scene, demonstrating its robustness and precision in adverse conditions (Fig. \ref{fig:real_experiments}(a)). Additionally, the detailed recording of the NTU Hive experiment is available at \url{https://www.youtube.com/watch?v=ocwUdYUv48s}, showcasing the system's performance in real-world applications. These results further highlight the adaptability and reliability of UA-MPC in complex, feature-sparse environments, making it a promising solution for advanced 3D mapping and localization tasks.

\begin{figure}
    \centering
    \includegraphics[width=0.48\textwidth]{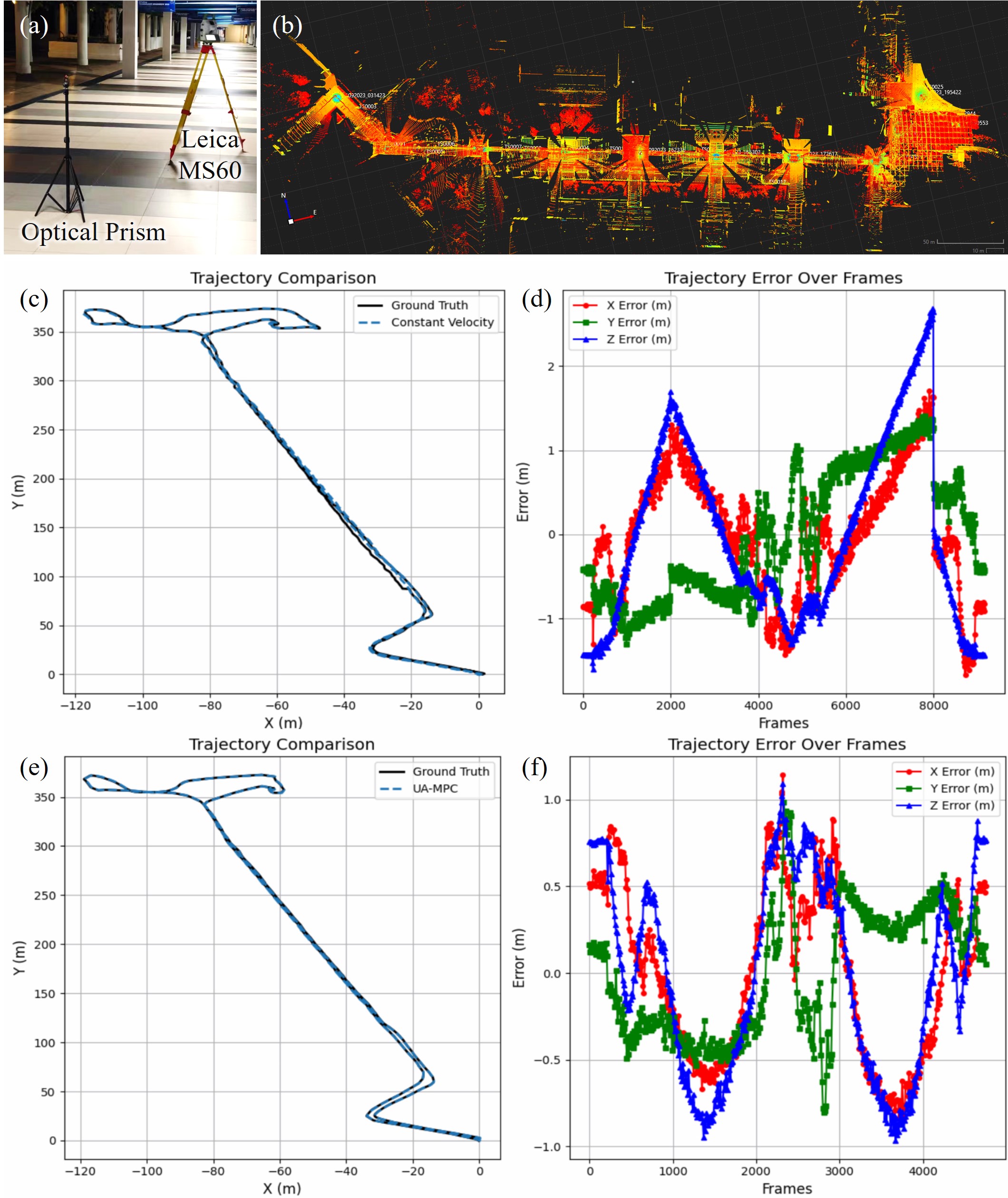}
    \caption{Absolute Translation Error (ATE) of the NTU Hive experiment. (a) Leica MS60 for Ground-Truth (GT) map; (b) GT map collected by the Leica MS60; (c) The trajectory comparison of GT trajectory and the motorized LiDAR system for constant-velocity control; (d) ATE of the motorized LiDAR system using constant-velocity control; (e) The trajectory comparison of GT trajectory and the motorized LiDAR system for UA-MPC; (f) ATE of the motorized LiDAR system using UA-MPC.}
    \label{fig:real_ate}
    \vspace{-0.5cm}
\end{figure}

Large-scale infrastructure, such as extensive corridors or tunnels, represents a critical class of facilities requiring regular monitoring for safety and maintenance purposes. To evaluate the proposed motorized LiDAR system's capabilities in such settings, we scanned the NTU Spine, a long corridor exceeding 300 meters in length, using our in-house motorized system. The scanning process is depicted in Fig. \ref{fig:real_experiments}(d), and the resulting 3D point cloud of the NTU Spine is shown in Fig. \ref{fig:real_experiments}(c). We try our best to collect the data using two different control strategies, namely, UA-MPC and constant velocity, following the same path to make a fair comparison. A detailed recording of the NTU Spine experiment is available at \url{https://www.youtube.com/watch?v=1H2dB0aJLSo}.
To assess the accuracy of the proposed motorized LiDAR system, we used the Leica M60 to collect a high-precision 3D map of the NTU Spine, as illustrated in Fig. \ref{fig:real_ate} (a) and (b). The Leica-generated map served as the ground truth (GT) reference. Using scan-to-map matching, we derived the GT trajectory for comparison with the LO trajectory generated by our system. The evaluations of the constant velocity control are shown in Fig. \ref{fig:real_ate}(c) and (d). The evaluations of the UA-MAP are shown in Fig. \ref{fig:real_ate}(e) and (f). 
The ATE of the constant velocity control is 1.51 m, while the ATE of the proposed motorized LiDAR system is measured at 0.80 m, demonstrating UA-MPC's high accuracy in large-scale 3D mapping tasks.

\section{Conclusion}

In this work, we address the challenge of achieving accurate and efficient 3D sensing using motorized LiDAR systems, which are critical for applications in photogrammetry and robotics. Existing systems often rely on fixed-speed motor controls, resulting in suboptimal performance in complex environments. To overcome this limitation, we introduced UA-MPC, an uncertainty-aware motor control strategy that optimally balances scanning accuracy and efficiency. By leveraging discrete observabilities predicted through ray tracing and modeling their distribution with a surrogate function, UA-MPC adapts motor control to varying scene complexities effectively. Additionally, we develop a ROS-based simulation environment tailored for motorized LiDAR systems, facilitating robust evaluation of motor control strategies across diverse scenarios. Extensive experiments on both simulated and real-world data demonstrated the capability of UA-MPC to enhance odometry accuracy without compromising scanning efficiency. Currently, the IMU and camera information is still not considered in the proposed system. In the near future, we will fully use the available sensing data including IMU and camera and explore the application of the motorized system on mobile robots.

\bibliographystyle{ieeetr}
\bibliography{ref} 

\end{document}